\documentclass[5pt]{article}

\usepackage[letterpaper]{geometry}
\usepackage{spconf,amsmath,epsfig}
\usepackage{enumitem}
\usepackage{multirow}
\usepackage{marvosym}
\usepackage{fancyhdr}
\begin{document}\sloppy

\def\x{{\mathbf x}}
\def\L{{\cal L}}

\title{SKELETON-BASED ACTION RECOGNITION USING LSTM AND CNN}
%
\name{Chuankun Li$^{1*}$, Pichao Wang$^{2*}$, Shuang Wang$^{1}$\Letter, Yonghong Hou$^{1}$, Wanqing Li$^{2}$}
\address{$^{1}$School of Electrical and Information Engineering, Tianjin University, China\\ $^{2}$Advanced Multimedia Research Lab, University of Wollongong, Australia\\ chuankunli@tju.edu.cn, pw212@uowmail.edu.au, wangshuang1993@tju.edu.cn, \\houroy@tju.edu.cn, wanqing@uow.edu.au}

%
%

\maketitle

\begin{abstract}
Recent methods based on 3D skeleton data have achieved outstanding performance due to its conciseness, robustness, and view-independent representation. With the development of deep learning, Convolutional Neural Networks (CNN) and Long Short Term Memory (LSTM)-based learning methods have achieved promising performance for action recognition. However, for CNN-based methods, it is inevitable to loss  temporal information when a sequence is encoded into images.  In order to capture as much spatial-temporal information as possible,  LSTM  and CNN are adopted to conduct effective recognition with later score fusion. In addition, experimental results show that the score fusion between CNN and LSTM performs better than that between LSTM and LSTM for the same feature. Our method achieved state-of-the-art results on NTU RGB+D datasets for 3D human action analysis. The proposed method achieved 87.40\% in terms of accuracy and ranked $1^{st}$ place in Large Scale 3D Human Activity Analysis Challenge in Depth Videos.
\end{abstract}
\begin{keywords}
3D Action Recognition, Convolutional Neural Networks, Long Short Term Memory
\end{keywords}
\renewcommand{\thefootnote}{}
\footnotetext{$*$ Equal contribution}
\footnotetext{\Letter~ Corresponding author}
\section{Introduction}
\label{sec:intro}
Human action recognition has attracted increasing attention from three types of input (RGB, depth and skeleton)~\cite{li2010action,pichao2015,pichaoTHMS,Pichaocvpr2017} in computer vision due to potential applications such as video search, intelligent surveillance systems and human-computer interaction. In the past few decades, RGB video data have been widely studied as input for human action recognition. However, compared to RGB data, depth modality is invariance to illumination and provides 3D structural information of the scene. With the advance of easy-to-use depth sensors and algorithms~\cite{shotton2011real} for highly accurate joint positions estimation from depth map, action recognition based on skeleton has been an active research topic. To date, many methods~\cite{chaudhry2013bio,hussein2013human,ohn2013joint,vemulapalli2014human,wang2012mining,zanfir2013moving,wang2014mining} based on handcrafted skeleton features have been reported for action recognition due to its conciseness, robustness and view-independent representation. However these features are shallow and dataset-dependent. Recently, deep learning methods have achieved outstanding performance in various computer vision tasks and several deep learning based methods~\cite{Pichao2016,hou2016skeleton,du2015hierarchical,zhu2016co,Songyang2017,liu2016spatio,Chuankun2017} for action recognition have been proposed. Currently, a common way to capture the spatio-temporal information in skeleton sequences is to use Convolutional Neural Networks (CNNs) or Recurrent Neural Networks (RNNs).

The challenge for CNN based method is how to effectively capture the spatio-temporal information of a skeleton sequence using image-based representation. In fact, it is inevitable to lose temporal information during the conversion of 3D information into 2D information. For example, Wang et al.~\cite{Pichao2016} encoded joint trajectories into texture images and utilized HSV (Hue, Saturation, Value) space to represent the temporal information. However, this method cannot distinguish some actions such as ``knock" and ``catch" due to the trajectory overlapping and losing past temporal information. Li et al.~\cite{Chuankun2017} proposed to encode the pair-wise distances of skeleton joints into texture images, and encoded the pair-wise distances between joints over a sequence of skeletons into color variations to capture temporal information. But this method cannot distinguish some actions of similar distance variations such as ``draw$-$circle$-$clockwise" and ``draw$-$circle$-$counter clockwise" due to the lost of local temporal information.

RNNs model the contextual dependency in the temporal domain, and have been successfully applied to processing sequential data with variable length such as language modeling and video analysis. Recently, several methods~\cite{du2015hierarchical,zhu2016co,Songyang2017} based on RNNs have been proposed for recognizing skeleton-based actions. The paper~\cite{Songyang2017} has proved that rich spatial domain features by exploring relative relations between joints perform better than original coordinates of joints. However, concatenation of different types of features cannot improve the performance compared with input with only a single type of feature. The reason is probably due to the  weak ability of RNNs in distinguishing useful information from multiple kinds of features.

In this paper, LSTM model is adopted to take advantage of its powerful ability to model the long term contextual information in the temporal domain. Different types of rich spatial domain features are fed to LSTM model. Inspired by the works ~\cite{Pichao2016,Chuankun2017,hou2016skeleton} where the authors fused class score from different planes,  in this paper class score fusion is adopted over different LSTM channels processing different types of features. In addition, score fusion is further conducted between CNN and LSTM channels. The way of this fusion method performs better than fusion between LSTM and LSTM for the same feature due to complementarity between CNN model and LSTM model. The proposed method was evaluated on NTU RGB+D Dataset and state-of-the-art results were achieved.

The rest of paper is organized as follows: Section 2 introduces the proposed method. Experimental results and discussions are presented in Section 3. Section 4 concludes the paper.

\section{Proposed Method }
The proposed method consists of three major components, as illustrated in Fig. 1, feature extraction from a skeleton sequence as the input of ten neural networks, including three LSTM models and seven CNN models, neural networks training and the late score fusion. Applying various methods on a skeleton sequence can obtain various features prominently in spatial or temporal domain, and we defined that features as SPF (spatial-domain-feature) and TPF (temporal-domain-feature). And we select SPF as the input of LSTM networks and TPF as the input of CNN networks. To obtain the SPF, three types of spatial domains features are extracted including R (relative position), J (distances between joints) and L (distances between joints and lines). To obtain the TPF, the methods used in~\cite{Chuankun2017} and~\cite{Pichao2016} are followed to generate the joint distances map (JDM) and joint trajectories map (JTM) respectively.
\begin{figure*}[t]
     \centering
\includegraphics[width=0.8\textwidth,height=0.38\textwidth]{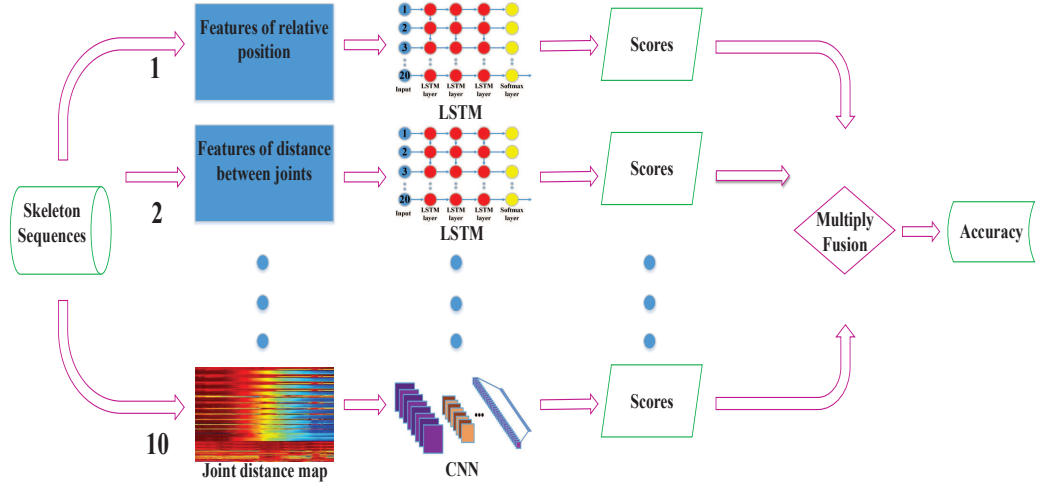}
\caption{The framework of the proposed method. The main composition of each route in the first three is Skeleton sequence-SPF-LSTM, and in the last seven routes is Skeleton sequence-TPF-CNN. \label{fig1}}
\end{figure*}
\subsection{Feature Extraction From Skeleton Sequence}
 The distance features are suitable to feed into LSTM due to rich spatial information, but the distance features lose a lot of orientation information. The features of the relative position is adopted to capture the orientation information. In this paper, we adopt a human skeleton model with 12 joints (hip center, spine, shoulder center, head, elbow left, wrist left, hand left, elbow right, wrist right, hand right, ankle left, and ankle right) due to their relative less noise. In the following, three types of skeletal features are described in detail.

 Let ${{p}_{j}}=(x,y,z)$ denote the 3D coordinate of the ${{j}^{th}}$ joint in each frame. The relative position descriptor $RP_{jk}^{t}$ and the Euclidean distance $JJD_{jk}^{t}$ between joints $j$ and joint $k$ at frame $t$ are denoted as
\begin{equation}
RP_{jk}^{t}=p_{j}^{t}-p_{k}^{t},j\ne k
\end{equation}
\begin{equation}
JJD_{jk}^{t}={{\left\| p_{j}^{t}-p_{k}^{t} \right\|}_{2}},j\ne k
\end{equation}
Compared to calculating the distance between joints, computing the distance between joints and lines has a higher complexity. Let ${{L}_{jk}}$ denote the line from joint $j$ to joint $k$. The distance $JL{{D}^{t}}(n,j,k)$ from joint $n$ to line ${{L}_{jk}}$ at the frame $t$ is expressed as follow:
\begin{equation}
JL{{D}^{t}}(n,j,k)=2S_{\Delta njk}^{t}/JJD_{jk}^{t},n\ne j\ne k
\end{equation}
where $S_{\Delta njk}^{t}$ is area of triangle formed by three joints(joint $n$, joint $j$ and joint $k$ ) and accelerated with Helen formula.
Any three joints form three lines. Thus, there are $3*C_{12}^{3}=660$ lines for one single subject with 12 joints. In fact, the number of features is extremely large when multiple subjects performing actions. For example, the 6072 lines will be generated for the interactions in NTU RGB+D Dataset. And it is very time consuming to use the large features to train the LSTM model. In order to reduce the computational cost, we follow the paper~\cite{Songyang2017} to select several important lines. And if one of the following three constraints is satisfied:
\begin{itemize}
  \item  The two joints are directly adjacent in the kinetic chain. This generates six lines.
  \item  If one joints is at the end of skeleton chain such as Hand joint, the other joint can be two steps away in the kinetic chain. This generates three lines.
  \item  The two joints are at end of skeleton chain. This produces ten lines.
\end{itemize}

For LSTM model, we enumerate three types of features with spatial information that are encoded in one pose and are independent of time. We follow the paper ~\cite{Pichao2016,Chuankun2017} to generate a color image and feed into CNN model. But the method in~\cite{Pichao2016} is view dependent, thus skeleton data is rotated around the $Y$ axis with a fixed step 45 degree to generate multi-view joint trajectory maps.
\subsection{Neural Networks }
In this paper, LSTM model and CNN model are adopted to process the different types of features. The two models are complementary to each other. LSTM can model the contextual dependency in the temporal domain very well. And CNN model can perform better in processing the image with more spatial information. Due to the complementation, the performance is greatly improved by the later score fusion.

Traditional RNNs only mines the short-term dynamics and are unable to discover relations among long-terms of inputs. In order to alleviate this drawback, a typical LSTM model is adopted. In this model, a unit contains four gates, an output state and an internal memory cell state. The LSTM transition equations are expressed as:
\begin{equation}
\left( \begin{array}{l}
{i_t}\\
{f_t}\\
{o_t}\\
{u_t}
\end{array} \right) = \left( \begin{array}{l}
\sigma \\
\sigma \\
\sigma \\
\tanh
\end{array} \right)\left( {W\left( \begin{array}{l}
{x_t}\\
{h_{t - 1}}
\end{array} \right)} \right)
\end{equation}
\begin{equation}
{c_t} = {i_t} \circ {u_t} + {f_t} \circ {c_{t - 1}}
\end{equation}
\begin{equation}
{h_t} = {o_t} \circ \tanh ({c_t})
\end{equation}
where ${i_t}$, ${f_t}$, ${o_t}$, ${u_t}$, ${h_t}$, ${c_t}$ are input gate, forget gate, output gate, input modulation gate, output state and internal memory cell state respectively. ${x_t}$ denotes the input to the network at time step $t$. Operator $\circ$ indicates element-wise product. The symbol $\sigma$ is the sigmoid function. Fig. 2 shows the schema of this recurrent unit.
\begin{figure}[t]
     \centering
\includegraphics[width=0.38\textwidth,height=0.22\textwidth]{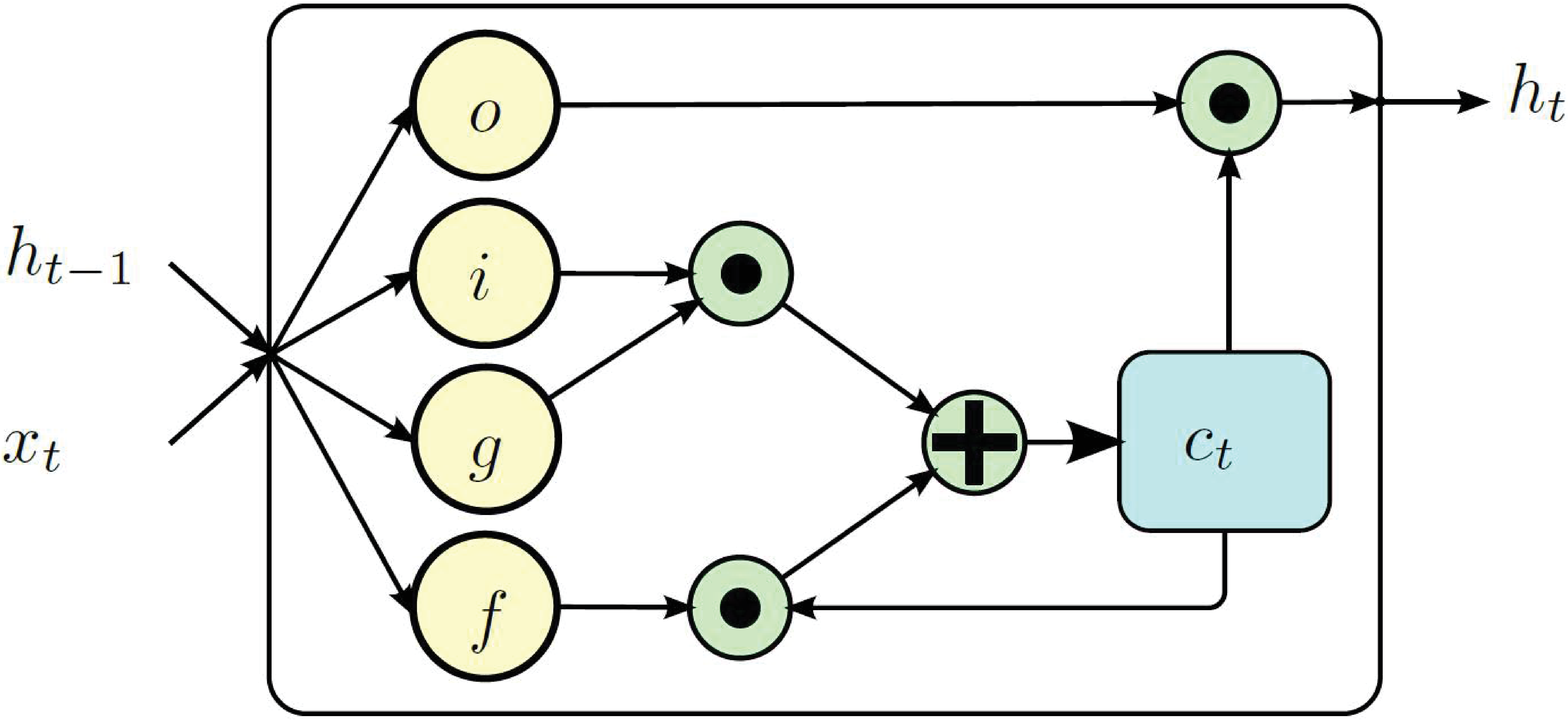}
\caption{Schema of a long short-term memory (LSTM) unit.\label{fig1}}
\end{figure}

As shown in Fig. 1, we build a 3-layer LSTM framework. The first LSTM layer takes the three types of features as the input ${x_t}$  and output ${h_t}$  of the lower LSTM layer is as the input ${x_t}$ of the upper LSTM layer. The output of the highest LSTM layer is fed to a softmax layer to transform the output codes to probability values of class labels.

 In this paper, the original AlexNet is adopted as CNN model. We follow the protocols used in~\cite{Pichao2016,Chuankun2017} to train the CNN model, namely fine-tuning on joint trajectory map and training from scratch on joint distances map. In fine-tuning, the pre-trained models on ILSVRC-2012 (Large Scale Visual Recognition Challenge 2012, a version of ImageNet) are used for initialization.
 \subsection{Score Fusion and Implementation Details  }
 \subsubsection{Score Fusion}
 Given a testing skeleton sequence, three feature vectors and seven texture maps are generated and fed into ten different trained neural networks. The score vectors of ten neural networks are fused and the max score in the resultant vector is assigned as the probability of the test sequence being the recognized class. The common fusion method consists of max-score fusion, average-score fusion and multiply-score fusion, that is, the score vectors outputted by the ten neural networks are maximized, averaged and multiplied in an element-wise way respectively. Compared with max-score fusion and average-score fusion, multiply-score fusion performs better and expressed as follows:
 \begin{equation}
 label = Fin(max({v_1} \circ {v_2} \cdot  \cdot  \cdot {v_9} \circ {v_{10}}))
 \end{equation}
 where $v$ is a score vector, $\circ$ refers to element-wise multiplication, and $Fin(\cdot)$ is a function to find the index of the element having the maximum score.
 \subsubsection{Implementation Details}
 Joint coordinates are normalized in a way similar to the scheme in~\cite{zanfir2013moving}, which imposes the same limbs (skeleton segments) lengths for poses obtained from different subjects. All 3D points are normalized based on average distances between adjacent joints from training data. The original point of body coordinate is translated to center hip joint. We also follow the paper~\cite{liu2016spatio} to divide a skeleton sequence to 20 sub-sequences and one frame is randomly selected from each sub-sequence.

 Caffe and Keras are as the deep learning platform and a NVIDIA TITAN X GPU card are adopted to run our experiments. We use three LSTM layers in the stacked network and the applied probability of dropout is 0.5. We train the network using stochastic gradient descent (SGD) for CNN model and Root Mean Square Propagation (RMSprop) for LSTM model. The learning rate was initially set to 0.001 for fine-tuning and 0.01 for the training from scratch on CNN model. And the learning rate was initially set to 0.001 and time step was set to 20 on LSTM model.
\section{Experiment Results }
The proposed method was evaluated on NTU RGB+D Dataset. Experiments were conducted to evaluate the effectiveness of three types of score fusion methods, score fusion of LSTM model in three types of features and score fusion between CNN model and LSTM model.
\subsection{NTU RGB+D Dataset}
To the best knowledge of authors, NTU RGB+D Dataset~\cite{shahroudy2016ntu}  is currently the largest dataset for action recognition. It consists of 56578 action samples of 60 different classes including front view, two side views, and left, right 45 degree views performed by 40 subjects aged between 10 and 35. This dataset is challenging and provides two types of evaluation protocols, cross-subject and cross-view.
\subsection{Evaluation of Fusion Methods}
The results of individual features and different fusion methods are listed in Table 1, where R, J, L, R-J-L concatenation, JTM, JDM represent relative position, distance between joints, distance between joints and lines, feature concatenation of above three features, joint trajectory map and joint distances map respectively. For example, JTM-xy stands for the joint trajectory map in xy plane. The scores of ten channels are fused by three types of methods including max-score fusion, average-score fusion and multiply-score fusion. And multiply-score fusion is adopted to evaluate the effectiveness of score fusion of LSTM model in three types of features (R, J, L) and score fusion between CNN model and LSTM model for the same features. For example, All-Max-Score fusion and R-J-L-Mul-Score fusion stand for the max-score fusion of ten channels and the multiply-score fusion of the three types of features respectively.

From Table 1, it can be seen that multiply-score fusion method improves the final accuracy substantially compared with average and max fusion methods. The concatenation of the three tpyes of features including relative position, distance between joints and distance between joints and lines cannot improve the performance well compared with a single type of feature, even performs worse than feature of distance between joints and lines in cross-view. The distance between joints is ultized to generate 3-D space joint distance map (JTM-xyz). JTM-xyz feeding to CNN model performs worse than distance between joints (J) feeding into LSTM, but JTM-xyz performs better when they fused with feature of relative position (R) feeding into LSTM model respectively. In addtion, it can be seen that the score fusion of LSTM and CNN greatly improved the final results.

\begin{table*}[t]
\begin{center}
\caption{Comparisons of the different schemes on the NTU RGB+D } \label{tab:cap}
\begin{tabular}{|c|c|c|}
  \hline
  \multirow{2}{*}{Techniques} & \multicolumn{2}{c|}{NTU RGB+D}\\
  \cline{2-3}
  &cross-subject&cross-view\\
  \cline{1-3}
  R (LSTM) & 70.98\%&75.97\%\\
  J (LSTM) & 67.62\%&77.95\%\\
  L (LSTM) & 69.16\%&80.63\%\\
  JTM-xy (CNN) & 70.22\%&73.89\%  \\
  JTM-xz (CNN) & 55.77\%&55.67\%\\
  JTM-yz (CNN)& 67.20\%&70.60\%\\
  JDM-xy (CNN)& 65.13\%&67.42\%\\
  JDM-xz (CNN)& 62.79\%&63.48\%\\
  JDM-yz (CNN)& 63.45\%&63.27\%\\
  JDM-xyz(CNN)& 66.79\%&74.90\%\\
  R-J-L concatenation (LSTM) & 71.26\%&80.00\%\\
  R-J-L-Mul-Score fusion (LSTM) & 76.52\%&85.35\%\\
  R-J-Mul-Score fusion (LSTM) & 74.50\%&82.19\%\\
  R-JDM-xyz-Mul-Score fusion (LSTM+CNN) & 76.06\%&83.05\%\\
 All-Max-Score fusion (LSTM+CNN)& 77.54\%&86.03\%\\
 All-Ave-Score fusion (LSTM+CNN)& 82.18\%&89.03\%\\
 All-Mul-Score fusion (LSTM+CNN)& \bfseries{82.89}\%&\bfseries{90.10}\%\\

  \hline
\end{tabular}
\end{center}
\end{table*}

\begin{table}[t]
\begin{center}
\caption{Experimental results (accuracies) on NTU RGB+D Dataset } \label{tab:cap}
\begin{tabular}{|c|c|c|}
  \hline
   Method&	Cross subject	&  Cross view\\
    \hline
  Lie Group~\cite{vemulapalli2014human}	&50.08\%	&52.76\%\\

Dynamic Skeletons~\cite{ohn2013joint}	&60.23\%	&65.22\%\\

HBRNN~\cite{du2015hierarchical}	&59.07\%	&63.97\%\\

Deep RNN~\cite{shahroudy2016ntu}	&56.29\%&	64.09\%\\

Part-aware LSTM~\cite{shahroudy2016ntu}&	62.93\%	&70.27\%\\

ST-LSTM+ Trust Gate~\cite{liu2016spatio}&	69.20\%	&77.70\%\\
JTM~\cite{Pichao2016} & 73.40\% & 75.20\%\\
Geometric Features~\cite{Songyang2017} &70.26\%	&82.39\%	\\
Clips+CNN+MTLN~\cite{ke2017new}  &79.57\% &84.83\%	\\
View invariant ~\cite{mengyuan2017}  &80.03\% &87.21\%	\\
Proposed Method	&\bfseries{82.89}\%&	\bfseries{90.10}\%\\
  \hline
\end{tabular}
\end{center}
\end{table}
From Table 2 we can see that our proposed method achieved the state-of-the-art results compared with both hand-crafted features based methods and deep learning methods. The method~\cite{vemulapalli2014human} is not good at modeling multi-person interactions. Dynamic Skeletons method~\cite{ohn2013joint} performed better than some RNN-based methods verifying the weakness of the RNNs~\cite{du2015hierarchical,shahroudy2016ntu}, which only mines the short-term dynamics and tends to overemphasize the temporal information even on large training data. LSTM with spatial-temporal domains~\cite{liu2016spatio} performs better than general LSTM~\cite{shahroudy2016ntu}, but it is worse than the Geometric Features with using conventional SLTM~\cite{Songyang2017}. Multi-CNN models~\cite{ke2017new,mengyuan2017} are widely used to extract the features and then fuse the features. The proposed method performs better than Multi-CNN models and achieves the best results in both cross-subject and cross-view evaluation.

\subsection{Result of Large Scale 3D Human Activity Analysis Challenge in Depth Videos}
The challange results are shown in Table 3.  From Table 3, it can be seen that the proposed method achieved $87.40\%$ in terms of accuracy and ranked the first in Large Scale 3D Human Activity Analysis Challenge in Depth Videos. The proposed method performs better than other methods which has verified the effectiveness of the proposed method.

\begin{table}[t]
\begin{center}
\caption{Result of Large Scale 3D Human Activity Analysis Challenge in Depth Videos } \label{tab:cap}
\begin{tabular}{|c|c|c|}
  \hline
   Rank&	Team Name&  Accuracy\\
    \hline
  1	&TJU (Our method)	&\bfseries{87.40}\%\\

2	&DIGCASIA	&85.14\%\\

3	&Inteam	&84.83\%\\

4	&404 NOT FOUND&	84.68\%\\

5&	Superlee	&84.01\%\\

6&	HRI	&83.80\%\\
7 & Knights & 81.23\%\\
8 &AMRL	&75.32\%	\\
9  &NJUST IMAG &75.06\%	\\
10  &XDUHZW&67.51\%	\\

  \hline
\end{tabular}
\end{center}
\end{table}

\section{Conclusions}
 In this paper, a simple yet effective method is proposed that based on LSTM and CNN score fusion. We take both advantages of LSTM and CNN models for final recognition: LSTM is good at exploiting strong temporal information while CNN is bias to mine strong spatial information. The effective multiply-score fusion method improved the recognition accuracy largely.  State-of-the-art results were achieved on  NTU RGB+D dataset and have verified the effectiveness of the proposed method.


\end{document}